\documentclass[runningheads,a4paper]{llncs}
\pdfoutput=1\usepackage{lmodern}\usepackage[utf8]{inputenc}
\usepackage{graphicx}
\usepackage{multirow}
\usepackage{enumitem}

\usepackage{hyperref}
\usepackage{xcolor}

\begin{document}
\title{Reading Comprehension in Czech via\\Machine Translation and Cross-lingual Transfer}
\titlerunning{Reading Comprehension in Czech via MT and Cross-lingual Transfer}
%
\author{Kateřina Macková\orcidID{0000-0001-9815-2763} \and Milan Straka\orcidID{0000-0003-3295-5576}}

\authorrunning{K. Macková, M. Straka}
%
\institute{Charles University, Faculty of Mathematics and Physics, \\
Malostranské náměstí 25, 118 00 Prague 1, Czech Republic \\
\email{\char`\{katerina.mackova@mff,straka@ufal.mff\char`\}.cuni.cz}
}

\maketitle              
\begin{abstract}
  Reading comprehension is a well studied
  task, with huge training datasets in English. This work focuses
  on building reading comprehension systems for Czech, without requiring
  any manually annotated Czech training data. First of all, we automatically
  translated SQuAD 1.1 and SQuAD 2.0 datasets to Czech to create training and development
  data, which we release at \url{http://hdl.handle.net/11234/1-3249}.
  We then trained and evaluated several BERT and XLM-RoBERTa baseline models. However, our main focus lies in cross-lingual transfer models.
  We report that a XLM-RoBERTa model trained on English data and evaluated
  on Czech achieves very competitive performance, only approximately 2 percent points
  worse than a~model trained on the translated Czech data. This result is extremely
  good, considering the fact that the model has not seen any Czech data during
  training. The cross-lingual transfer approach is very flexible and provides
  a reading comprehension in any language, for which we have enough monolingual raw texts.

\end{abstract}

\section{Introduction}

The goal of a reading comprehension system is to understand given text
and return answers in response to questions about the text. In English, there exist
many datasets for this task, some of them very large. In this work, we consider
the frequently used SQuAD 1.1 dataset~\cite{SQuAD1}, an English
reading comprehension dataset with around 100,000 question-answer pairs, which is
widely used to train many different models with relatively good accuracy.
We also utilize SQuAD 2.0 dataset~\cite{SQuAD2}, which combines SQuAD 1.1 dataset
with 50,000 unanswerable questions linked to already existing paragraphs,
making this dataset more challenging for reading comprehension systems.

In this paper, we pursue construction of a reading comprehension system for Czech
without having any manually annotated Czech training data, by reusing English
models and English datasets. Our contributions are:

\begin{itemize}
    \item We translated both SQuAD 1.1 and SQuAD 2.0 to Czech by
      state-of-the-art machine translation system~\cite{popel-2018-nmt}
      and located the answers in the translated text using MorphoDiTa~\cite{MorphoDiTa2014} and DeriNet~\cite{vidra-etal-2019-derinet}, and released the dataset.
    \item We trained several baseline systems using BERT and XLM-RoBERTa architectures,
      notably a system trained on the translated Czech data, and a~system which
      first translates a text and a question to English, uses an English model, and
      translates the answer back to Czech.
    \item We train and evaluate cross-lingual systems based on BERT and XLM-RoBERTa,
      which are trained on English and then evaluated directly on Czech. We report
      that such systems have very strong performance despite not using any Czech data
      nor Czech translation systems.
    \end{itemize}

\section{Related Work}

There exist many English datasets for reading comprehension and question answering,
the readers are referred for example to~\cite{SQuAD1} for a nice overview.

Currently, the best models for solving reading comprehension are based
on BERT architecture~\cite{BERT} (which is a method of unsupervised pre-training
of contextualized word embeddings from raw texts), or on some follow-up
models like ALBERT~\cite{ALBERT} or RoBERTa~\cite{RoBERTa}.

Most BERT-like models are trained on English, with two notable exceptions.
Multilingual BERT (mBERT), released by~\cite{BERT}, is a single language model
pre-trained on monolingual corpora in 104 languages including Czech;
XLM-RoBERTa (XLM-R)~\cite{XLM-RoBERTa} is a similar model pre-trained on
100 languages, and is available in both \textit{base} and \textit{large}
sizes, while only \textit{base} mBERT is available.

Cross-lingual transfer capability of mBERT has been mentioned in 2019 by many
authors, for example by Kondratyuk et al.~\cite{kondratyuk-straka-2019-75}
for morphosyntactic analysis or by Hsu et al.~\cite{ZeroShotRC} for reading comprehension.

Very similar to our paper is the parallel independent work of Lewis et al.~\cite{MLQA},
who perform cross-lingual transfer evaluation of reading comprehension models
on six non-English languages (neither of them being Czech).

\section{Constructing Czech Reading Comprehension Dataset}

The SQuAD 1.1 dataset consists of 23,215 paragraphs belonging to 536 articles.
Attached to every paragraph is a set of questions, each with several possible
answers, resulting in more than 100,000 questions. While the train and the
development datasets are public, the test set is hidden. We refer the readers
to~\cite{SQuAD1} for details about the dataset construction,
types of answers and reasoning required to answer the questions.

The SQuAD 2.0 dataset~\cite{SQuAD2} extends SQuAD 1.1 with more than 50,000
unanswerable questions linked to the existing paragraphs.

\subsection{Translating the Data and Locating the Answers}
\label{sec:translating_data}

We employed the English-Czech state-of-the-art machine translation system~\cite{popel-2018-nmt}
to translate the SQuAD data.\footnote{Available on-line at {\small\url{https://lindat.mff.cuni.cz/services/translation/}}.}  Translation of all texts,
questions and answers of SQuAD 2.0 took 3 days.

Because the answers are subsequences of the given text in SQuAD, we also
needed to locate the translated answers in the text. We considered several alternatives:
\begin{itemize}
  \item Estimate the alignment of the source and target tokens using attention of the machine
  translation system, and choose the words aligned to the source answer. Unfortunately,
  we could not reliably extract alignment from the attention heads of a Transformer-based
  machine translation system.
  \item Mark the answer in the text before the translation, using for example quotation marks, similarly to~\cite{MLQA}. Such an approach would however result in a~dataset
  with every question linked to a custom text, which would deviate from the SQuAD
  structure.
  \item Locate the answer in the given text after the translation, without relying on
  the assistance from the machine translation system.
\end{itemize}

We chose the third alternative and located the translated answers in the texts as follows:
\begin{enumerate}
  \item We lemmatized the translated text and answer using MorphoDiTa~\cite{MorphoDiTa2014}.
  \item We replaced the lemmas by roots of their word-formation relation trees according
  to the DeriNet 2.0 lexicon~\cite{vidra-etal-2019-derinet}.
  \item Then we found all continuous subsequences of the text with the same DeriNet roots
  as the answer, but with any word order.
  \item Finally, if several occurrences were located, we chose the one with the relative
  position in the text being the most similar to the relative position of the original answer
  in the original text.
\end{enumerate}

We believe the proposed algorithm has substantially high precision (after manually verifying
many of the located answers),
and we also find its recall satisfactory. Notably, in the SQuAD 2.0 training dataset,
we have preserved 107,088 questions (which is 82.2\% of the English ones)
and in the development dataset we kept 10,845 questions, 91.3\% of the original dataset.
The detailed sizes of the created Czech datasets are presented in Table~\ref{tab:pres-trans}.
Note that the ratio of the kept data in SQuAD 1.1 is lower,
because unanswerable questions of SQuAD 2.0 are always preserved.

The dataset is available for download at \url{http://hdl.handle.net/11234/1-3249}.

\begin{table}
    \caption{Size of the translated Czech variant of SQuAD 1.1 and SQuAD 2.0.}
    \label{tab:pres-trans}
    \centering
    \setlength{\tabcolsep}{10pt}
    \begin{tabular}{|l|l|l|l|l|}
        \hline
        \multicolumn{2}{|c|}{\multirow{2}{*}{\bfseries Dataset}} & {\bfseries English} & {\bfseries Czech} & {\bfseries Percentage}\\
        \multicolumn{2}{|c|}{} & \textbf{Questions} & \textbf{Questions} & \textbf{Kept}\\\hline\hline
        \multirow{2}{*}{SQuAD 1.1} & Train & 87,599 & 64,164 & 73.2\%\\\cline{2-5}
        & Test & 10,570 & 8,739 & 82.7\%\\\hline\hline
        \multirow{2}{*}{SQuAD 2.0} & Train & 130,319 & 107,088 & 82.2\%\\\cline{2-5}
        & Test & 11,873 & 10,845 & 91.3\%\\\hline
    \end{tabular}
\end{table}


\subsection{Evaluation Metrics}

The SQuAD dataset is usually evaluated using two metrics: \textbf{exact match},
which is the accuracy of exactly predicted answers, and \textbf{F1-score}
computed over individual words of the answers.

Given that Czech is a morphologically rich language, we performed lemmatization
and then replaced lemmas by DeriNet roots (as in Section~\ref{sec:translating_data})
prior to evaluation with the official evaluation script.


\section{Model Training and Evaluation}

Considering that the current best SQuAD models are all BERT based, we
also employ a BERT-like architecture. We refer readers to~\cite{BERT} for
detailed description of the model and the fine-tuning phase.

Because our main goal is Czech reading comprehension, we consider
such BERT models which included Czech in their pre-training, notably
Multilingual BERT (mBERT), released by~\cite{BERT}, both cased and uncased,
and also XLM-RoBERTa (XLM-R)~\cite{XLM-RoBERTa}, both \textit{base} and \textit{large}.
As a reference, we also include English BERT \textit{base}, both cased and uncased.

We finetuned all models using the \raisebox{-3pt}{\includegraphics[width=12pt]{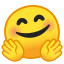}} \texttt{transformers} library \cite{huggingface-transformer}. For all \textit{base} models, we used
two training epochs, learning rate 2e-5 with linear warm-up of 256 steps and batch size 16;
for XLM-RoBERTa we increased batch size to 32 and for XLM-RoBERTa \textit{large}
we decreased learning rate to 1.5e-5 and increased warm-up to 500.

\begin{figure}[ht!]
  \centering
  \includegraphics[width=.83\hsize]{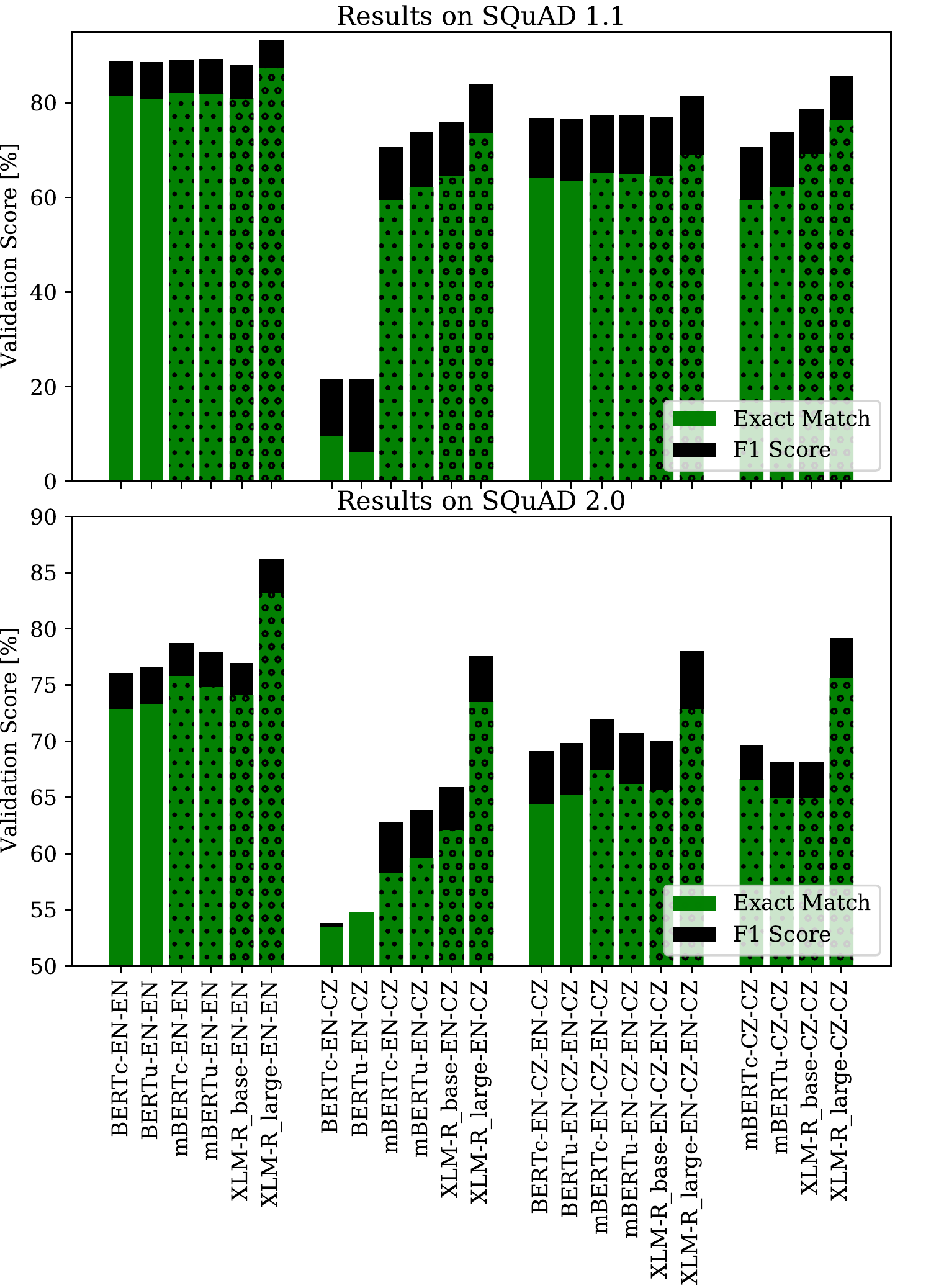}
  \caption{Development set performance of all models for English and Czech SQuAD 1.1
  and SQuAD 2.0 datasets.}
  \label{fig:1.1-2.0all}
\end{figure}

All our results are presented in Table~\ref{tab:1.1-2.0all} and also graphically in Figure~\ref{fig:1.1-2.0all}.

\begin{table}[ht]
  \caption{Development performance of English and Czech models on SQuAD 1.1, 2.0.}
  \label{tab:1.1-2.0all}
  \centering
  \setlength{\tabcolsep}{6pt}
  \begin{tabular}{|l|l|l||r|r||r|r|}\hline
    \multicolumn{1}{|c|}{\multirow{2}{*}{\bf Model}} &
      \multicolumn{1}{c|}{\multirow{2}{*}{\bf Train}} &
      \multicolumn{1}{c||}{\multirow{2}{*}{\bf Dev}} &
      \multicolumn{2}{c||}{\bf SQuAD 1.1} &
      \multicolumn{2}{c|}{\bf SQuAD 2.0} \\\cline{4-7}
    & & & \multicolumn{1}{c|}{\bf EM} & \multicolumn{1}{c||}{\bf F1} &
          \multicolumn{1}{c|}{\bf EM} & \multicolumn{1}{c|}{\bf F1} \\\hline
    BERT cased    & EN & EN       & 81.43\% & 88.88\% & 72.85\% & 76.03\% \\\hline
    BERT uncased  & EN & EN       & 80.92\% & 88.59\% & 73.35\% & 76.59\% \\\hline
    mBERT cased   & EN & EN       & 81.99\% & 89.10\% & 75.79\% & 78.76\% \\\hline
    mBERT uncased & EN & EN       & 81.98\% & 89.27\% & 74.88\% & 77.98\% \\\hline
    XLM-R base    & EN & EN       & 80.91\% & 88.11\% & 74.07\% & 76.97\% \\\hline
    XLM-R large   & EN & EN       & 87.27\% & 93.24\% & 83.21\% & 86.23\% \\\hline
    \hline
    BERT cased    & EN & CZ       & 9.53\%  & 21.62\% & 53.48\% & 53.84\% \\\hline
    BERT uncased  & EN & CZ       & 6.16\%  & 21.75\% & 54.78\% & 54.83\% \\\hline
    mBERT cased   & EN & CZ       & 59.49\% & 70.62\% & 58.28\% & 62.76\% \\\hline
    mBERT uncased & EN & CZ       & 62.09\% & 73.89\% & 59.59\% & 63.89\% \\\hline
    XLM-R base    & EN & CZ       & 64.63\% & 75.85\% & 62.09\% & 65.93\% \\\hline
    XLM-R large   & EN & CZ       & 73.64\% & 84.07\% & 73.50\% & 77.58\% \\\hline
    \hline
    BERT cased    & EN & CZ-EN-CZ & 64.06\% & 76.78\% & 64.35\% & 69.11\% \\\hline
    BERT uncased  & EN & CZ-EN-CZ & 63.57\% & 76.61\% & 65.26\% & 69.86\% \\\hline
    mBERT cased   & EN & CZ-EN-CZ & 65.09\% & 77.47\% & 67.40\% & 71.96\% \\\hline
    mBERT uncased & EN & CZ-EN-CZ & 65.00\% & 77.38\% & 66.20\% & 70.72\% \\\hline
    XLM-R base    & EN & CZ-EN-CZ & 64.52\% & 76.91\% & 65.62\% & 70.00\% \\\hline
    XLM-R large   & EN & CZ-EN-CZ & 69.04\% & 81.33\% & 72.82\% & 78.04\% \\\hline
    \hline
    mBERT cased   & CZ & CZ       & 59.49\% & 70.62\% & 66.60\% & 69.61\% \\\hline
    mBERT uncased & CZ & CZ       & 62.11\% & 73.94\% & 64.96\% & 68.14\% \\\hline
    XLM-R base    & CZ & CZ       & 69.18\% & 78.71\% & 64.98\% & 68.15\% \\\hline
    XLM-R large   & CZ & CZ       & 76.39\% & 85.62\% & 75.57\% & 79.19\% \\\hline
  \end{tabular}
\end{table}

\subsubsection{English}
For reference, we trained and evaluated all above models on English SQuAD 1.1 and SQuAD 2.0.
The results are consistent with the published results. It is worth noting that the only
\textit{large} model reaches considerably better performance, and that mBERT achieves
better results than English BERT.

\subsubsection{Czech Training, Czech Evaluation}
Our first baseline model is trained directly on the Czech training data and
then evaluated on the development set. The relative performance of the BERT
variants is very similar to English, but the absolute performance is considerably lower.
Several facts could contribute to the performance decrease -- a smaller training
set, noise introduced by the translation system and morphological richness of the
Czech language.

\subsubsection{English Models, Czech Evaluation via Machine Translation}

Our second baseline system (denoted CS-EN-CS in the results) reuses English models to
perform Czech reading comprehension -- the Czech development set is first translated
to English, the answers are then generated using English models, and finally translated
back to Czech.

The translation-based approach has slightly higher performance for \textit{base}
models, which may be caused by the smaller size of
the Czech training data. However, for the \textit{large} model, the direct approach
seems more beneficial.

\subsubsection{Cross-lingual Transfer Models}

The most interesting experiment is the cross-lingual transfer of the English models,
evaluated directly on Czech (without using any Czech data for training).
Astonishingly, the results are very competitive with the other models evaluated
on Czech, especially for XLM-R \textit{large}, where there are within 1.6 percent
points in F1 score and 2.75 percent points in exact match of the best Czech model.

\subsection{Main Findings}

\subsubsection{Why Does Cross-lingual Transfer Work}
The performance of the cross-lingual transfer model is striking.
Even if the model never saw any Czech reading comprehension data
and it never saw any parallel Czech-English data, it reaches nearly
the best results among all evaluated models.

This strong performance is an indication that mBERT and XLM-R represent different
languages in the same shared space, without getting an explicit
training signal in form of parallel data. Instead, we hypothesise that
if there is a large-enough similarity among languages, the model
exploits it by reusing the same part of the network to handle this phenomenon
across multiple languages. This in turn saves capacity of the model and allows
reaching higher likelihood, improving the quality of the model. In other words,
greedy decrease of a loss function performed by SGD is good enough motivation
for representing similarities in a~shared way across languages.

Furthermore, word embeddings for different languages demonstrate a remarkable
amount of similarity even after a simple linear transformation, as demonstrated
for example by \cite{artetxe-etal-2018-robust} or \cite{conneau-etal-2017-word-translation}.
Such similarities are definitely
exploitable (and as indicated by the results also exploited) by BERT-like models
to achieve shared representation of multiple languages.

\subsubsection{Pre-training on Czech is Required}
The strong performance of cross-lingual models does not necessarily mean
the models can ``understand'' Czech -- the named entities could be similar enough
in Czech and English, and the model could be capable of answering without
understanding the question.

Therefore, we also considered an English reading comprehension model based
on English BERT, which did not encounter any other language but English during
pre-training. Evaluating such a model directly on Czech delivers
surprisingly good performance on SQuAD 2.0 -- the model is unexpectedly good
in recognizing unanswerable questions. However, the performance of such model
on SQuAD 1.1 is rudimentary -- 9.53\% exact match and 21.62\% F1-score, compared
to 62.90\% exact match and 73.89\% F1-score of an mBERT uncased model.


\subsubsection{Cased versus Uncased}
Consistently with intuition, cased models seem to perform generally better than
uncased. However, in the context of cross-lingual transfer, we repeatedly
observed uncased models surpassing the cased ones. We hypothesise that this result
could be caused by larger intersection of Czech and English subwords
of the uncased models (which discard not only casing information, but also diacritical marks),
because larger shared vocabulary could make the cross-lingual transfer easier.







\section{Conclusion}
In this paper, we have explored Czech reading comprehension without
any manually annotated Czech training data. We trained several baseline
BERT-like models using translated data, but most importantly we evaluated
a cross-lingual transfer model trained on English and then evaluated directly
on Czech. The performance of this model is exceptionally good, despite the fact that
no Czech training data nor Czech translation system was needed to train it.

\section*{Acknowledgements}

The work was supported by the Grant Agency of the Czech Republic,
project EXPRO LUSyD (GX20-16819X) and by the SVV 260 575 grant of Charles University.
This research has also been using data and services provided by the
LINDAT/CLARIAH-CZ Research Infrastructure (\url{https://lindat.cz}), supported by the
Ministry of Education, Youth and Sports of the Czech Republic (Project No. LM2018101).

\bibliographystyle{splncs04}
\bibliography{tsd20_czech_qa}

\end{document}